\pdfoutput=1

\documentclass[11pt]{article}

\usepackage[]{acl}

\usepackage{times}
\usepackage{latexsym}

\usepackage[T1]{fontenc}

\usepackage[utf8]{inputenc}

\usepackage{microtype}

\usepackage{hyperref}
\usepackage{url}

\usepackage{booktabs}
\usepackage{graphicx}
\graphicspath{{figures/}}
\usepackage{caption}
\usepackage{subcaption}
\usepackage{enumitem}

\newcommand{\cready}[1]{} 
\usepackage{algorithm,algorithmic}
\usepackage{threeparttable}
\usepackage{xcolor,colortbl}
\definecolor{LightGray}{rgb}{0.7,0.7,0.7}
\usepackage{multirow}
\usepackage{placeins}
\usepackage{wrapfig}
\usepackage{amsfonts}

\title{On Neurons Invariant to Sentence Structural Changes\\in Neural Machine Translation}

\author{Gal Patel\quad\quad Leshem Choshen\quad\quad Omri Abend \\
  School of Computer Science and Engineering\\
  The Hebrew University of Jerusalem \\
  \texttt{first.last@mail.huji.ac.il} }

\begin{document}
\maketitle
\begin{abstract}
We present a methodology that explores how sentence structure is reflected in neural representations of machine translation systems. 
We demonstrate our model-agnostic approach with the Transformer English-German translation model.
We analyze neuron-level correlation of activations between paraphrases while discussing the methodology challenges and the need for confound analysis to isolate the effects of shallow cues. We find that similarity between activation patterns can be mostly accounted for by similarity in word choice and sentence length.
Following that, we manipulate neuron activations to control the syntactic form of the output. We show this intervention to be somewhat successful, indicating that deep models capture sentence-structure distinctions, despite finding no such indication at the neuron level.
To conduct our experiments, we develop a semi-automatic method to generate meaning-preserving minimal pair paraphrases (active-passive voice and adverbial clause-noun phrase) and compile a corpus of such pairs.\footnote{The dataset is provided with paraphrase generation code: \url{https://github.com/galpatel/minimal-paraphrases}}
\end{abstract}

\section{Introduction}


Understanding the roles neurons play is important for the interpretability of neural machine translation (NMT) models. Finding neurons that are either invariant or sensitive to particular structural distinctions may explain how such structures are encoded, and validates the robustness of translation systems, which is a challenging but important problem \citep{10.1145/3377811.3380339, freitag-etal-2020-human}. Furthermore, understating how these encodings are used by the network may potentially enable controlling the output by direct manipulation of neurons.


Previous work analyzing what aspects of sentence structure are encoded in network representations mostly took a probing approach or focused on syntactic agreements.
Works that compared activations did so either across models with identical input \citep{dalvi2019one, bau2018identifying, wu-etal-2020-similarity} or by representation words, not sentences \citep{DBLP:journals/corr/abs-2110-07483}. Our novelty lies in using the same model with paraphrased input pairs, to analyze sentence structure encoding, without probing (c.f. \S\ref{sec:related_work}). We derive inspiration from Computer Vision works that analyze model behavior under non-semantic changes to the input \citep{7298701, NIPS2009_428fca9b}.

For our proposed approach, we provide a dataset of minimal paraphrases (along with a code to extend it). We take two phenomena as test cases: active to passive voice and an adverbial clause to a noun phrase (see Table \ref{tab:data_examples} and section \S\ref{sec:dataset}). 

\begin{table*}[ht]
\begin{small}
\begin{center}
\begin{tabular}{@{}lll@{}}
\toprule & Source & Paraphrased \\ \midrule
\begin{tabular}[c]{@{}l@{}}Active Voice$\rightarrow$ Passive Voice\end{tabular}   & \textit{She \textbf{took} the book}                                                                & \textit{The book \textbf{was taken} by her}                                                        \vspace{0.1cm}\\
\begin{tabular}[c]{@{}l@{}}Adverbial Clause$\rightarrow$ Noun Phrase\end{tabular} & \textit{\begin{tabular}[c]{@{}l@{}}The party died down before \textbf{she arrived}\end{tabular}} & \textit{\begin{tabular}[c]{@{}l@{}}The party died down before \textbf{her arrival}\end{tabular}} \\ \bottomrule
\end{tabular}
\end{center}
\vspace{-0.25cm}
\caption{Examples produced by our paraphrasing engine}
\label{tab:data_examples}
\end{small}
\end{table*}


To compare the activation patterns of sentences that may be comprised of a different number of tokens, we aggregate tokens representations. We then measure the correlation of neuron activations between paraphrases and provide a confound analysis. We find that the main contributors to strong correlation are similar positional encodings and bag-of-words overlap, suggesting strong correlation is derived from similar input encoding and not high-level abstractions learned by the model. The identification of these confounds may be beneficial to future work on network analysis (see \S\ref{sec:correlation}).\looseness=-1

Our findings suggest that these paraphrase distinctions are nonetheless encoded and used, as evident by our experiments of manipulating neurons (\S\ref{sec:manipulation}). We control the structure of the translation output (e.g., active/passive) by translating neuron activations in a fixed direction. We show that this manipulation generates outputs that are more similar to the desired form with an in-depth evaluation, using BLEU, dependency parsing, and manual analysis. We provide ablation studies that show that the results are not simply random artifacts of manipulation, with a non-local effect. Lastly, we compare different methods for selecting subsets of neurons to be manipulated (\S\ref{sec:man_tb}) with counter-intuitive results, attributed to neurons of general importance and multiple roles per neuron.

Overall, the similarity between neuron activation over paraphrases is mostly explained by shallow input features: the positional and token embeddings. Therefore, some neurons represent input features, but sentence-level information is not localized, even in higher layers. Moreover, we show that sentence phrasing can be na\"ively controlled, with a manipulation of a large number of neurons. This suggests that the distinction between different sentence structures is encoded in the model, probably in a distributed manner. Lastly, the neurons most effective for such manipulations are the ones most correlated across paraphrases, not necessarily those that vary the most.

\section{Dataset: Minimal Paraphrase Pairs}\label{sec:dataset}

We curate datasets that isolate specific sentence structure distinctions. To achieve that, we require sentence pairs with the following attributes:
\begin{itemize}[noitemsep,leftmargin=*]
    \item\textbf{Similar Meaning}, to have invariant semantics.
    \item \textbf{Minimal Change}, to facilitate the experimental setup and the interpretation of the results.
    \item \textbf{Controlled Change}, where paraphrasing is consistent and well-defined. As opposed to lexical paraphrases that tend to be idiosyncratic, the same distinction is applied to all instances. \item \textbf{Reference Translation}, since we examine translation models.
\end{itemize}

Existing paraphrasing tools and datasets fail to satisfy these criteria (see \S\ref{sec:related_work}). Therefore, we develop our own paraphrasing method, with which we compile two parallel sets: active voice to passive voice and an adverbial clause to a noun phrase. Sentence examples can be found in Table~\ref{tab:data_examples}.

The proposed process is automatic, following predefined syntactic rules while utilizing several NLP models. First, we identify sentences that match some source patterns (active voice, adverbial clause) according to a Dependency Parsing and POS tags model \citep{spacy} and a Semantic Role Labeling model \citep{gardner-etal-2018-allennlp}. Then, we rephrase the sentence to the desired structure.
We complement missing prepositions by choosing the one with the highest probability as predicted by BERT \citep{devlin-etal-2019-bert}. For example, the adverbial clause sentence \textit{``She felt accomplished when she met the investor''} requires the preposition "\textit{with}" in the noun phrase form \textit{``She felt accomplished during her meeting} with  \textit{the investor''}, and the temporal preposition \textit{when} is replaced with \textit{during}. In ambiguous instances, we choose whether or not to insert a preposition by opting for the sentence with the higher probability according to GPT2 Language Model \citep{radford2019language}.
When replacing a verb with a noun (e.g., \textit{arrival} is replaced with \textit{arrive}), we look for the most suitable conversion in existing lexicons, including Nomlex \citep{Macleod98nomlex:a}, \href{https://amr.isi.edu/download/lists/morph-verbalization-v1.01.txt}{AMR's} and \href{https://github.com/monolithpl/verb.forms.dictionary}{Verb Forms}. See Appendix \ref{ap:data} for details and examples.\footnote{The dataset is provided with paraphrase generation code: \url{https://github.com/galpatel/minimal-paraphrases}}

\begin{table}[tbh]
\begin{small}
\begin{center}
\begin{tabular}{@{}lll@{}}
\toprule
                 & Paraphrased & Valid \\ \midrule
Adverbial Clause to Noun Phrase & 376       & 114   \\
Active Voice to Passive Voice    & 3107      & 1169  \\ \bottomrule
\end{tabular}
\caption{Minimal Paraphrase pairs count, as derived from WMT19 English-German dev set, before (left) and after (right) filtering.}
\label{tab:pairs_stats}
\vspace{-0.25cm}
\end{center}
\end{small}
\end{table}
We apply our paraphrasing engine to the WMT19 English-German development set \citep{barrault2019findings}.
Some results are disfluent. For example, the sentence \textit{``He took his time''} is converted to \textit{``His time was taken by him''}, which is syntactically well-formed, but anomalous. Therefore, we manually filter the data. For more details and other failed filtering approaches, see Appendix \ref{ap:filtering}. The number of pairs is given in  Table~\ref{tab:pairs_stats}.

\section{Technical Setup}

\paragraph{Model.} We demonstrate our model-agnostic methodology with the Transformer model for Machine Translation \citep{NIPS2017_3f5ee243}. We use the fairseq implementation \citep{ott2019fairseq}, which was trained on the WMT19 English-German train set \citep{barrault2019findings}. The embedding dimension is 1024 with sinusoidal positional encoding. 


\paragraph{Dataset.} Minimal paraphrases (see~\S\ref{sec:dataset}). Due to space considerations, we present results on the active/passive set in the main paper. Clause/noun phrase results are in appendices \S\ref{appendix:corr_clause} and \S\ref{appendix:manipulation}.

\paragraph{Notations.} We refer to trained models with a different random seed as $m_1, m_2$. 
We denote the set of source sentences $S=\{s_1, ..., s_n\}$, and its corresponding paraphrased set with $P=\{p_1, ..., p_n\}$ (e.g., $s_i$ is an active voice sentence and $p_i$ is its passive counterpart). We follow \citet{liu-etal-2019-linguistic, wu-etal-2020-similarity} and take into account only the last sub-word token for each word (results with all sub-word tokens were similar). We consider as neurons the 1024 activation values in the output embedding of the 6 encoder layer blocks \citep[following][]{wu-etal-2020-similarity}\footnote{Experiments on activations internal to each layer block have similar but weaker effects (Appendix \ref{appendix:inside_block}).}.
This leads to the following definition:
the activation of some neuron $l$ in model $m$, on sentence $s_i$ is $x^{m, l}_S[i]$, while $x^{m, l}_S$ is a vector of size $n$ (with $n$ being the number of sentences in set $S$, i.e. one activation value per sentence, see \S\ref{sec:aggregate}).

\setlength\abovedisplayskip{3pt}
\setlength\belowdisplayskip{3pt}

\section{Detecting Correlation Patterns}\label{sec:correlation}
To detect activation patterns under experimental conditions, we measure Pearson correlation\footnote{ Results with Spearman correlation were similar.} between neural activations. 
While correlation analysis has been previously used to analyze neuron-level behavior \citep{bau2018identifying, dalvi2019one, wu-etal-2020-similarity, Meftah2021NeuralSD}, our novelty lies with the independent variable being a property of the input (i.e., paraphrases) and not the model (e.g., architecture, initialization).

\subsection{Sample Alignment Challenge}\label{sec:aggregate}

For every neuron, we measure its activation values while feeding a model with either the set of source sentences as input (e.g., active voice) or the paraphrased set (e.g., passive voice).
Since the number of words may differ between paired sentences, so will the total number of tokens in these sentence sets. Consequently, the neural activation sample size will vary, which poses a challenge for testing correlation.
Previous works did not face this difficulty, as they compared activation values given the same input corpus. 


We overcome this with intra-sentence aggregation. Instead of having an activation value per word in the input corpus, we consider a single value per sentence, by pooling activations of words within a sentence. Ideally, this will allow for a sentence-level analysis of semantics and structure. Mean pooling was previously considered in several instances. \citet{ethayarajh-2019-contextual} compared sentences by averaging their word vectors and \citet{DBLP:journals/corr/abs-2110-07483} aggregated words with specified attributes by averaging their representation, element-wise. The main results of this paper use mean pooling. In Appendix \ref{appendix:pooling} we also report results with min/max pooling, that show similar trends.

It is possible to consider other straightforward approaches, which we find less suitable.
One option is a position-wise alignment of words 
(discarding the last words of the longer sentence). The difficulty here is that words of different semantics and syntactic roles are compared. For example, the third words of the active/passive sentence in Table~\ref{tab:data_examples} would be $the$ and $was$, the former is a determiner of a direct object while the latter is an auxiliary of the root verb.
Another option is functional correspondence alignment, where we measure correlation only between the functional tokens that indicate the structure change (e.g. $took$ vs. $taken$ and $arrived$ vs. $arrival$ from the examples in Table~\ref{tab:data_examples}). That would result in an analysis based on similar single words with the same context words but in different syntactic forms. This could be problematic as it would capture a local syntactic change but not necessarily a sentence-level phrasing.

\subsection{Baseline Experiments}

We capture correlation across paraphrases, denoted with \textit{ParaCorr}. Given the same model, we look at activations over a set of sentences and their correlation to the activations over the paraphrased set:
\noindent
\begin{equation}\label{eq:2}
ParaCorr(l, l')=\rho(x^{m_1, l}_S, x^{m_1, l'}_P)
\end{equation}
\noindent

ParaCorr should examine how neural networks represent differences in sentence structure (or similarities in semantics).
We follow by comparing it to \textit{ModelCorr} - the correlation between any pair of neurons across models, when given the same input:
\noindent
\begin{equation}\label{eq:1}
ModelCorr(l, l')=\rho(x^{m_1, l}_S, x^{m_2, l'}_S)
\end{equation}
\noindent
ModelCorr is based on \citet{bau2018identifying} who detected generally important neurons in this way.

\begin{figure*}[tb]
\begin{center}
    \begin{subfigure}{.24\textwidth}
      \includegraphics[width=1\linewidth]{heat/data_active_11.png}
      \caption{ParaCorr}
      \label{fig:para_corr}
    \end{subfigure}
    \begin{subfigure}{.24\textwidth}
      \includegraphics[width=1\linewidth]{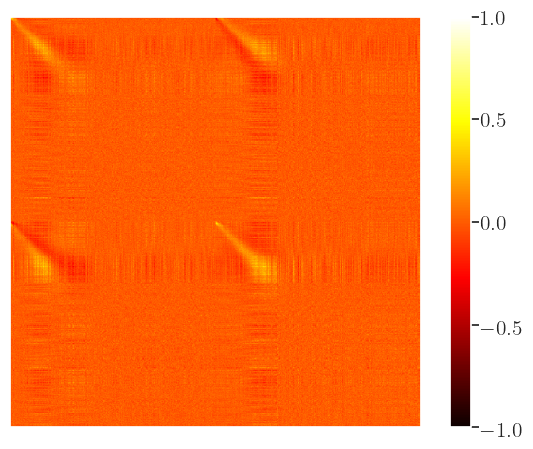}
      \caption{ModelCorr}
      \label{fig:model_corr}
    \end{subfigure}
    \begin{subfigure}{.24\textwidth}
      \includegraphics[width=1\linewidth]{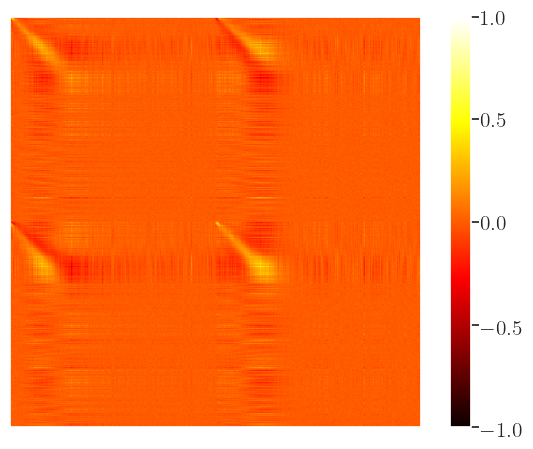}
      \caption{PosCorr}
      \label{fig:pos_corr}
    \end{subfigure}
    \begin{subfigure}{.24\textwidth}
      \includegraphics[width=1\linewidth]{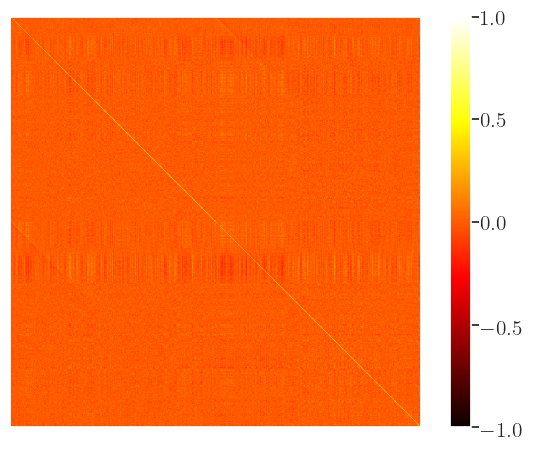}
      \caption{TokenCorr}
      \label{fig:token_corr}
    \end{subfigure}
    \end{center}
    \caption{Activation correlation of neurons in the encoder, using the active-passive dataset.}
    \label{fig:corr_maps}
\end{figure*}

Figures \ref{fig:para_corr} and \ref{fig:model_corr} show ParaCorr and ModelCorr correlation maps.\footnote{We feature only the first layer due to resolution constraints. Any effect shown is present in all layer block pairs but weakens when moving away from the main diagonal (i.e. correlation across layers) or when the layers are higher.}
Some of ParaCorr's observed effect also appears in ModelCorr, suggesting the observed correlations might be unrelated to the examined variable, i.e. paraphrases. Moreover, ModelCorr indicates a strong correlation between neurons of the same location in different models, but we had expected a different pattern as highly correlated neurons should be distributed differently for randomly initialized models.

\subsection{Controlling for Confounds}\label{sec:conf}
In this section, we show that strong activation correlations between paraphrases are a product of low-level cues. Namely, we inspect how the propagation of token identity and positional information greatly influences the correlation. This is a relevant confound to note for previous work adapting correlation analysis on neurons \citep{bau2018identifying, wu-etal-2020-similarity, Meftah2021NeuralSD}. The positional encoding in our setting is sinusoidal, therefore the same positions are encoded exactly the same across models. Paraphrases present a minor change in sentence length: $2.0
\pm{0.4}$ or $0.8\pm{0.8}$ token difference when paraphrasing active to passive or clause to noun phrase, respectively. The positional encodings are therefore similar. As for tokens, paraphrases have a large unigram overlap.

We define \textit{PosCorr} as activation correlation between sentences with identical positional encoding but different token embeddings.
Formally:
\noindent
\begin{equation}\label{eq:3}
PosCorr(l, l')=\rho(x^{m_1, l}_S, x^{m_1, l'}_{\hat{S}})
\end{equation}

\noindent
Where $\hat{S}$ is a set of random token sequences, uniformly sampled from the dictionary, with the same sequence lengths as $S$.
PosCorr isolates the strong correlation effect observed both in ModelCorr and ParaCorr (Figure \ref{fig:pos_corr}). Repetition through the layers is probably due to the residual connections, which propagate the positional encoding. Indeed, when we looked at correlations of neurons inside the layer block -- before the first residual connection -- the effect seen in PosCorr was missing (see Appendix \ref{appendix:inside_block}). The implication is that input representation, and not higher-level learned representation, is likely the cause of strong correlations.

\textit{TokenCorr} accounts for token embeddings. We strip an input set $S$ from its positional encoding, denoted by $\tilde{S}$, and measure correlation:
\noindent
\begin{equation}\label{eq:4}
TokenCorr(l, l')=\rho(x^{m_1, l}_S, x^{m_1, l'}_{\tilde{S}})
\end{equation}
\noindent
TokenCorr (Figure \ref{fig:token_corr}) captures the diagonals phenomenon of ParaCorr, explained by paraphrases having a large bag-of-words overlap (the effect is not present in ModelCorr since token embeddings are different across models). This implies that individual token identities, and not necessarily sentence-level semantics, contribute to strong correlations. This distinction is made apparent when we consider how word order may affect meaning. For example, \textit{"Rose likes Josh"} has a different meaning than \textit{"Josh likes Rose"}, although the sentences have the same bag of words. Even if word meaning is sufficient for a lot of cases, grammatical cues are still essential \citep{DBLP:journals/corr/abs-2201-12911}.

We further dissect the observed correlation for possible confounds. First, we compare activations of sentence pairs that share only the relevant syntactic structure (e.g., two random active voice sentences). No strong correlation is observed (between -0.17 to 0.20). This suggests that the effect observed in the TokenCorr experiment, where the same tokens are fed to the model (Figure \ref{fig:token_corr}, Eq.~\ref{eq:4}) is not explained by a similar structure cue (i.e., active voice). In another experiment, we combine both PosCorr and TokenCorr: we strip the original sentence from its positional embedding and replace the tokens with random ones -- i.e., nothing is shared between the compared conditions. As little correlation is detected (between -0.27 to 0.31), we rule out the possibility of neurons with constant values. 

Overall, our confound analysis implies the following:
(1) strong activation correlation is greatly due to low-level components and not high-level learned knowledge, (2) strong correlation detected across paraphrases may not be exclusive to sentences with similar meaning but different structures, and (3) sentence structure is not localized to a specific set of neurons in our analysis.

\section{Manipulation of Neurons}\label{sec:manipulation}

Manipulation of neurons allows us to control the output translation (without additional training) and adds a causative explanation to the role neurons play.
We look into changing the activation values to force the output to have a desired syntactic structural feature (e.g., active or passive voice). Although we did not observe individual neurons that have a strong positive/negative correlation across paraphrases in \S\ref{sec:correlation}, these sentence-structure distinctions could still be encoded in a decentralized manner in the model, and therefore susceptible to manipulation.
We address three main questions:
\begin{enumerate}[nosep]
  \item Can we effectively control the output structural properties by changing neuron values?
  \item Does the exact activation value matter or only the identity of the modified neurons?
  \item How to choose a set of neurons to manipulate?
\end{enumerate}

\subsection{Setup}

We denote with $\bar{x_c}[i]$ the average activation of neuron \textit{i} given a set of input sentences with property \textit{c}, e.g., the property of passive voice. For a model with a total of $n$ neurons (in the encoder), we have a vector of average behavior $\bar{x_c}\in{R}^n$. 
An intervention on neuron $i$ with a current value of $x[i]$ from property $c_1$ towards property $c_2$ is a linear translation between their averages:
\noindent
\begin{equation}\label{eq:manipulation}
\hat{x}[i]=x[i]-\beta(\bar{x_{c_1}}-\bar{x_{c_2}})[i]
\end{equation}
\noindent

So far, the formulation is based on previous work \citep{bau2018identifying}. Our preliminary experiments showed it is essential that the scaling factor $\beta$ includes a normalization term because comparing the effects of different manipulations (i.e. different target properties $c_2$) can be confounded by activation magnitude. Therefore, $\beta=\frac{\alpha}{\|{\bar{x_{c_1}}-\bar{x_{c_2}}}\|}$. This leaves us with various parameters to experiment with: the properties $c_1, c_2$ we wish to manipulate, the subset of neurons we intervene with, and the scaling factor $\alpha$. We explore the former two in \S\ref{sec:man_exp} and the latter in Appendix \ref{appendix:magnitude}, having $\alpha=1$ as default.

We evaluate whether manipulation increases the similarity of the output to a reference with the target form ($c_2$), relative to the similarity with the source form ($c_1$).
We measure BLEU scores between our model's translation and Google translations, which (in the absence of manual references) we consider as references to both the source and target forms. This is a reasonable assumption given the performance gap between the models we use and Google Translate. Later we discuss evaluation by additional methods to complement BLEU (see \S\ref{sec:beyond_bleu}).

\subsection{Experiments}\label{sec:man_exp}

\begin{figure*}[tbh]
    \begin{center}
    \begin{subfigure}{.32\textwidth}
      \includegraphics[width=1\linewidth]{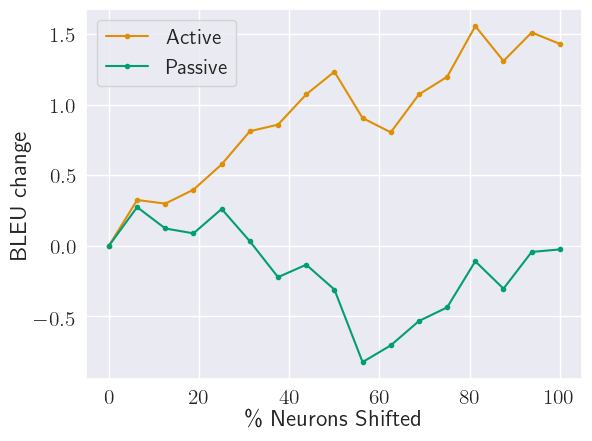}
      \caption{Controlled baseline}
      \label{fig:passive_baseline}
    \end{subfigure}
    \begin{subfigure}{.32\textwidth}
      \includegraphics[width=1\linewidth]{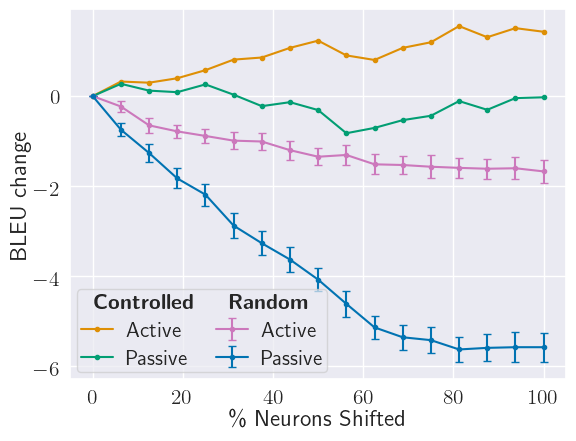}
      \caption{Random direction comparison}
      \label{fig:passive_randdir}
    \end{subfigure}
    \begin{subfigure}{.32\textwidth}
      \includegraphics[width=1\linewidth]{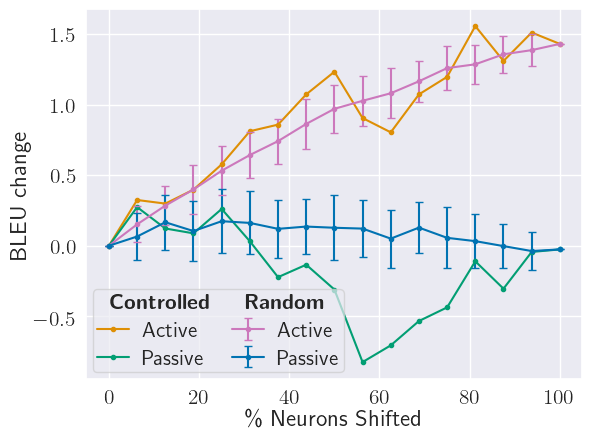}
      \caption{Random selection comparison}
      \label{fig:passive_randselect}
    \end{subfigure}
    \end{center}
    \caption{Manipulating the outputted translation to be of active voice when feeding passive voice input. Lines present BLEU change with active and passive references, as a function of the amount of neurons manipulated (x-axis). 
    For random experiments (b) and (c) we report the average of the measured BLEU and its standard deviation.}
    \label{fig:ap_shift_bleu}
\end{figure*}

We present experiments on manipulating passive voice inputs toward active voice translations. The reverse manipulation (active input to passive translation) and the results on the clause/noun-phrase set can be found in Appendix \ref{appendix:manipulation}.

\paragraph{Baseline Manipulation.} 
We modify an increasing amount of neurons, first selecting the neurons most correlated across paraphrases (i.e., we rank by $ParaCorr(l,l)$ with the higher values first). The motivation to use the correlation as a rank is based on \citet{bau2018identifying}, who used it as an indicator of important neurons.
We manipulate passive voice inputs toward active voice translations. Outputs become more similar to active voice than passive voice (Figure \ref{fig:passive_baseline}), suggesting that sentence structure is indeed encoded in the model, even if we did not detect the distinction at the neuron-level in \S\ref{sec:correlation}. Moreover, the information is \textit{used} by the model when generating translations and it can be controlled.

\paragraph{Direction of Manipulation.}
We explore the importance of the manipulation direction, i.e. the value we shift towards. A random manipulation, with a random vector $y_r\in\mathbb{R}^n$, is defined by:
\noindent
\begin{equation}\label{eq:rand}
\hat{x}[i]=x[i]-\beta(\bar{x_{c_1}}-y_r)[i]
\end{equation}
\noindent
Repeating 100 different random vectors (Figure \ref{fig:passive_randdir}), we find it to be substantially worse. This implies that success is tied to the specific values we manipulate, not an artifact of any modification.

\paragraph{Selection of Manipulation.}
 We test whether there is a preferable subset of neurons to manipulate by randomly selecting a subset of neurons.\footnote{We also experimented with choosing random neurons under the constraint they have the same distribution among the 6 encoder layers as the controlled case, with similar results.} Results (Figure \ref{fig:passive_randselect}) do not indicate that a controlled selection of neurons (according to ParaCorr ranking) is better than random. Overall, it seems that a large subset of neurons has to be modified to obtain the desired outcome, which agrees with our correlation results, where the active-passive feature was not localized. Notwithstanding, we study subset choice even further in \S\ref{sec:man_tb}.

\subsection{Beyond BLEU}\label{sec:beyond_bleu}

BLEU score captures translation quality on the surface and not necessarily how good (or bad) it is at capturing form (active vs. passive). Therefore, we employ additional evaluation measures.
\paragraph{Passive Score.} Specifically for the active-passive dataset, we use a dependency parser and a POS tagger to detect passive form\footnote{Using Spacy \citep{spacy}, we consider a sentence to be in passive voice if the root lemmatization is \textit{"werden"} and it has a child of dependency \textit{"oc"} (i.e., clausal object) with a tag indicating a participle form.}. The scorer is not intended to be perfect in capturing all passive instances\footnote{Limited recall: baseline translation of passive sentences (without manipulation) gets a score of 37.38\%}, but it could serve as a complementary measure to indicate trends. We observe a decrease of detected passive voice when we manipulate the passive input towards active translation (see Figure \ref{fig:tb_passive}), solidifying the BLEU results.

\paragraph{Qualitative Analysis.} A native German speaker examined a sample of output translations and found successful manipulations (see Appendix \ref{ap:quality}). She discussed failed outputs -- where the translation changed (i.e. unequal strings) but did not result in the desired form. They did not degrade the translation. In some cases, sentences changed between stative passive and dynamic passive, rather than between passive and active (the distinction between these passive types is more evident in German). In other cases, the manipulation was not applicable. For instance, some verbs could not be in an adverbial form in German, which demands them either to appear as a noun phrase or to be replaced with a synonym verb (an example is in Appendix \ref{ap:quality}). These suggest that the manipulation has successfully modified the desired attributes in the sentence, even when not automatically detected as such. Moreover, it may be limited by the nature of the target language and the model's capabilities to generalize to synonyms while controlling the sentence structure.

\paragraph{Held-out Test Set.}
We repeated the manipulation experiment on a held-out test set: 552 active voice sentences from the WMT19 test set. This allows us to examine if the successful manipulation effect extends to a setting where the manipulated sentences do not contribute to the measure of average activation of the source form.
As can be seen in \ref{ap:test_set}, the manipulation still results in the desired change in passive form detection.

\paragraph{Linearity Caveat.} As \citet{ravfogel-etal-2021-counterfactual} noted, positive results can indicate a causal effect, while negative results should be interpreted carefully since we have a linear manipulation in a non-linear setting. We leave the exploration of non-linear techniques for future work.

\section{Specific Neuron Set Selection}\label{sec:man_tb}
\begin{figure}[tb]
    \begin{center}
    \begin{subfigure}{.22\textwidth}
      \includegraphics[width=1\linewidth]{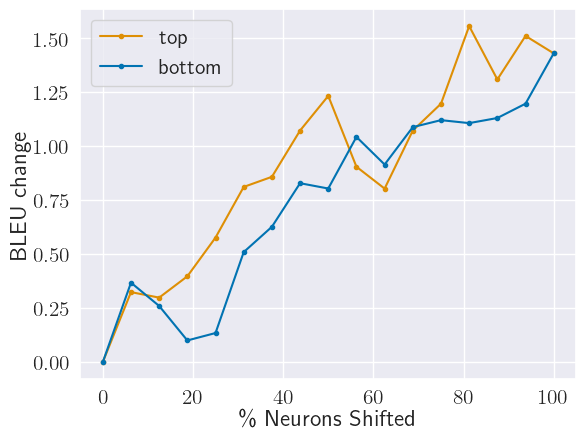}
      \caption{BLEU}
      \label{fig:tb_bleu}
    \end{subfigure}
    \begin{subfigure}{.22\textwidth}
      \includegraphics[width=1\linewidth]{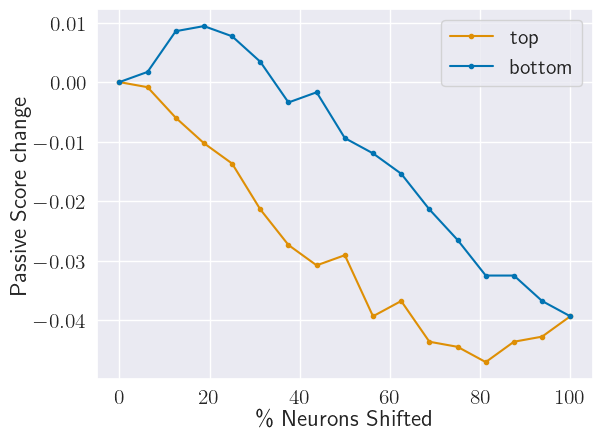}
      \caption{Passive Score}
      \label{fig:tb_passive}
    \end{subfigure}

    \end{center}
    \caption{Top ParaCorr neurons are better for manipulation. Manipulating the output translation to be in active voice when feeding passive voice as input. Comparing the choice of neurons to manipulate when starting from the top or bottom according to the rank given by ParaCorr. (1) Measuring by BLEU against active voice references and (2) measuring passive score that automatically detects passive voice.}
    \label{fig:top_bottom}
\end{figure}

\begin{figure}[tbh]
    \begin{center}
      \includegraphics[width=.32\textwidth]{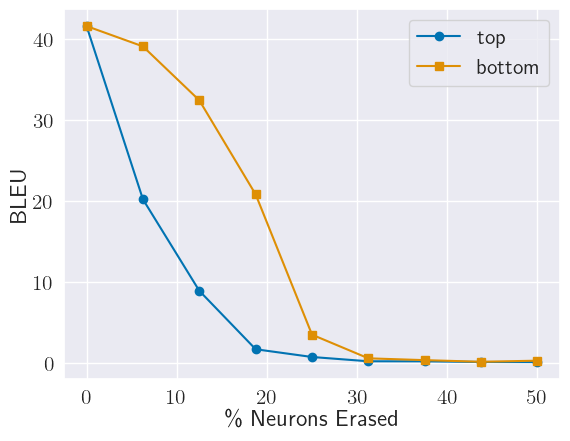}
    \end{center}
    \caption{Top ranked neurons have a stronger impact on the translation quality of a test set, measured in BLEU. Erasure of neurons from the top or bottom of the rank given by the value of correlation between paraphrases. }
    \label{fig:erase}
\end{figure}

Although finding a subset of neurons that carries a specific functionality is a difficult problem \citep{Sajjad2021NeuronlevelIO}, we look for subsets that are \textit{relatively} better for manipulation. In our baseline (\S\ref{sec:man_exp}) we chose which neurons to manipulate according to the rank given by ParaCorr (i.e., sorting neurons by $ParaCorr(l,l)$, high to low).
Under an intuitive interpretation, neurons that positively correlate when a systematic change is made to the input are those invariant to that change. Neurons with a negative correlation are specific to the change in sentence structure. Following these, we expect bottom-ranked neurons to be better for manipulation than top-ranked neurons. Contrarily, we observe the opposite phenomenon (Figure \ref{fig:top_bottom}). The following tests may explain it.

\paragraph{Model Performance.} Top-ranked neurons are important for overall performance. We follow \citet{bau2018identifying} who identified important neurons by deleting them (i.e., setting activations to zero) and examining the impact on the model performance. We delete an increasing amount of neurons, according to ParaCorr rank. We measure BLEU on a held-out set of 552 active voice sentences, and their references, extracted from the WMT19 test set. Results (Figure \ref{fig:erase}) show that top-ranked neurons have a stronger impact on the translation quality than bottom-ranked do, suggesting that ParaCorr partially ranks neurons by their general importance. 

\begin{figure}[tbh]
    \begin{center}
     \includegraphics[width=.32\textwidth]{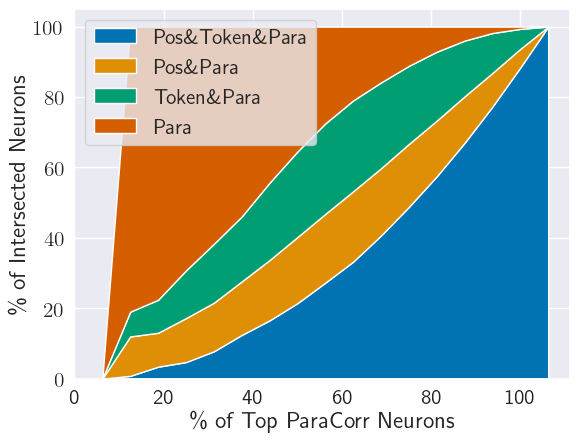}
    \end{center}
    \caption{Top ParaCorr neurons intersect with neurons most related to token embeddings and positional encoding. The x-axis represents the amount of top ParaCorr neurons as a percentage of all the neurons in the encoder. The y-axis shows how many of these \textit{x} top ParaCorr neurons are also in the set of top \textit{x} TokenCorr neurons and/or top \textit{x} PosCorr neurons. The y-axis scale is a percentage out of \textit{x}. Measured on the active-passive set. }
    \label{fig:overlap}
\end{figure}

\paragraph{Role Overlap.} Some of the top ParaCorr neurons account for lexical identity and positional information. This fact explains why they have the most impact when manipulating sentence structure. Word order is essential for active-passive paraphrasing, where the subject and direct object exchange places. Notably, when we tested non-paired active-passive sentences the phenomena did not repeat itself, see Appendix \ref{appendix:splitdata}. Word tokens are the building blocks for the semantic meaning of a sentence (which is the same across paraphrases), even when a bag-of-word is not exclusive to a single meaning.
The first evidence to support this claim is seen in \S\ref{sec:correlation}, where most of the strong correlations in ParaCorr are explained by similarity in the tokens and the positional embeddings between the inputs (i.e., TokenCorr and PosCorr, respectively). In an additional test, we check how many of the top ParaCorr neurons are also in the top PosCorr and TokenCorr neurons. Figure \ref{fig:overlap} shows that for any count \textit{x}, the set of top \textit{x} ParaCorr neurons have an intersection with the sets of top \textit{x} PosCorr or TokenCorr neurons.

\vspace{-0.25cm}

\section{Related Work}\label{sec:related_work}

\paragraph{Understanding NLP Neural Networks.} Various approaches were previously proposed \citep{belinkov-glass-2019-analysis}, each with a methodology that differs from ours.
Probing tasks investigate whether linguistic properties of the input text can be effectively predicted from model representations \citep{jawahar-etal-2019-bert, tenney-etal-2019-bert, slobodkin2021mediators}. They shed light on what information is kept within a model, but not necessarily what is used, or how \citep{DBLP:journals/corr/abs-2110-07483}. Others employ mediation analysis theory: \citet{DBLP:journals/corr/abs-2004-12265, finlayson-etal-2021-causal} study semantic behavior and syntactic agreement, respectively.
Some works analyze attention heads \citep{voita-etal-2019-analyzing} or follow attention flow \citep{abnar-zuidema-2020-quantifying}.
Visualization tools interpret activations, but with some exceptions \citep[e.g.,][]{7298701}, they are mostly limited to qualitative examples. \citet{durrani-etal-2020-analyzing} interpreted individual neurons with probing-like methods for fine-grained analysis.
Challenge sets \citep{choshen-abend-2019-automatically, 10.1162/tacl_a_00321} and adversarial examples \citep{alzantot-etal-2018-generating}, expose challenging cases by analyzing the NMT system's behavior, rather than representation.
\citet{DBLP:journals/corr/abs-2102-01017}
analyzes semantically equivalent inputs by clustering their embeddings. They improve prediction by continual training, while we manipulate translation post-training.

\paragraph{Interpretation in other domains.}
Some Computer Vision work inspired our approach. \citet{7298701} study the interaction between input transformation and its representation through the layers, while \citet{NIPS2009_428fca9b} examine invariant neurons, those that are selective to high-level features but robust given semantically identical transformations. Their methodologies do not fit the NLP domain since they rely on a mathematically well-defined input transformation (e.g., rotation). We propose an alternative with our paraphrases in \S\ref{sec:dataset}. 
Linguistic encoding in the human brain has been studied in neuroscience works. \citet{Friederici2011TheBB} analyzed the correlation of neuroimaging where subjects are presented with sentences with subtle syntactic variations or violations, and found that well-correlated regions are considered to process syntax. \citet{FedorenkoE6256} presented human subjects with various inputs, which are analogous to our correlation experiments: word lists (TokenCorr), meaningless grammatical sentences (PosCorr), non-words lists (combination of TokenCorr and PosCorr), and regular sentences (ParaCorr). 

\paragraph{Individual neurons analysis with correlation.} \citet{bau2018identifying} detected neurons that correlate across LSTM models while showing these are the most important for performance. They manipulated individual neurons to control single words in the output (e.g., gender, tense), with their linguistic role identified by probing with a GMM classifier. The technique to identify neurons that behave similarly in different models was previously suggested by \citet{dalvi2019one}, who found neurons in LSTM models to have role polysemy, aligning with our discussion in \S\ref{sec:man_tb}. Later, \citet{wu-etal-2020-similarity} employed correlation to examine similarities of different Transformer architectures. \citet{Meftah2021NeuralSD} adapted correlations to quantify the impact of fine-tuning by measuring activations of neurons before and after domain adaptation.

\paragraph{Controlling active-passive voice in translation.}
\citet{Yamagishi2016ControllingTV} controlled voice (active/passive) in RNN-based machine translation, from Japanese to English, when an indicator was given as input. Their method required additional model training, unlike ours.

\paragraph{Paraphrases.}
Existing paraphrasing tools vary by how localized their edits are.
Some alter the lexical level \citep{ribeiro-etal-2018-semantically}, other alter whole phrases \citep{ganitkevitch-2013-large,10.1145/1597735.1597764}, some are sentence-level paraphrases \citep{dolan2004unsupervised}, while some split source sentences into sub-sentences \citep[e.g.,][]{dornescu-etal-2014-relative,lee-don-2017-splitting}.
Other than paraphrasing tools, existing datasets include the PPDB database \citep{pavlick-etal-2015-ppdb} that contains sentence paraphrases that are lexical, phrasal, or syntactic.
\citet{DBLP:journals/corr/abs-1904-01130, dolan2005automatically} include paraphrase and non-paraphrase pairs, the former with high lexical overlap, while \citep{Hu_Rudinger_Post_Van_Durme_2019} contains multiple paraphrases of lexical diversity. None of these match our criteria for paraphrases (\S\ref{sec:dataset}).

\section{Conclusion}

With our curated dataset, we introduced a model-agnostic methodology to detect activation patterns across paraphrases. By a meticulous confound analysis, we found that activation similarity is likely due to shallow features of sequence length or word identity, which are not exclusive to meaning-preserving variations. We emphasize how these confounds must be taken into account when attempting to detect local correlation under any experimental setup. 
We controlled syntactic structures of generated output, which provides evidence of the ability of models to capture them. While we found the modification technique to be important for manipulation success, the selection of a subset of neurons was more challenging. 
Future work should test additional architectures and language pairs or examine the representation significance of our paraphrase pairs in other NLP tasks.

\section*{Limitations}\label{ap:limits}
Our work has some limitations. First, we compile the minimal paraphrases dataset to analyze representations of an isolated difference. Even so, human language is complex and any input transformation could not be mathematically well-defined. Our paraphrases may have other differences than the ones we indicate, which may introduce noise to our analysis. For example, we don't know the effect specific verbs (e.g., more common/rare ones) have when they appear in noun form, or what possible bias 'by' introduces when we add it for passive voice.

Secondly, we demonstrate our model-agnostic methodology in a specific setting with a transformer model for en-de translation. Our insights of what is captured (or isn't) may change when experimenting on other architectures or language pairs.

We aggregate token representations to sentence representation, discussing our choice and other possible approaches in \S\ref{sec:aggregate}. We do this to overcome potential differences in the number of tokens the paraphrases contain. However, the aggregation may lose encoded information along the way. We find some evidence of that when examining other pooling techniques (min/max) in Appendix \ref{appendix:pooling}.


Our manipulation shows that many neurons have to be modified for a successful outcome (at least ~50\%). Still, when manipulating more and more neurons, the effect is not always monotone in every setting (see Figure \ref{fig:ap_shift_bleu_active}, Figure \ref{fig:nphrase2clause_shift_bleu} and Figure \ref{fig:clause2nphrase_shift_bleu}). We conducted our qualitative analysis on the output with all of the encoder neurons modified and see positive results, with a discussion of how to choose neurons relatively better (\S\ref{sec:man_tb}).

In the following Appendix sections we also address: the filtering required for the automatically generated paraphrases (\S\ref{ap:filtering}), analysis on neurons internal to the layer block (\S\ref{appendix:inside_block}), possible alternative explanation for the observed results for top vs. bottom ranked neurons (\S\ref{appendix:splitdata}), unsuccessful manipulation cases (\S\ref{sec:clause2nphrase_diss}) and manipulation magnitude (\S\ref{appendix:magnitude}). 

\section*{Acknowledgements}

This work was supported by the Israel Science Foundation (grant no. 2424/21). We thank Yonatan Belinkov for helpful comments and discussion. We thank Nicole Gruber for her work on paraphrase annotation and qualitative analysis of German translations.

\bibliography{anthology,custom}
\bibliographystyle{acl_natbib}

\appendix

\section{Compilation of Minimal Paraphrase Pairs}\label{ap:data}

\subsection{Tools and Techniques}\label{sec:tools}
We explain, in greater detail, the main tools we use when we paraphrase, as briefly discussed in section \S\ref{sec:dataset}.
\paragraph{Pattern Detection.}
Making sure we change form but not semantics, we rely on syntax-based patterns, not word-based. We use dependency parsing (including Part of Speech tagging) and Semantic Role Labeling combined (by \citet{spacy} and \citet{gardner-etal-2018-allennlp}, respectively) to detect active form and adverbial clauses by type (see Table \ref{tab:clause_para}).

\paragraph{Sentence Probability}\label{tool:sen_prob} Used for choosing between two sentence options (with or without a certain preposition). We use gpt2 model \citet{radford2019language} by huggingface \citet{wolf-etal-2020-transformers} to get sentence probability for each option and opt for the higher.

\paragraph{Word Insertion.} 
\begin{algorithmic}
\STATE{\bfseries Input:} a sentence $X=x_1,x_2,..., x_n$ , a position \textit{i}, and a set of possible words\\$W=\{w_1, ..., w_m\}$
\end{algorithmic}
\begin{algorithmic}[1]
		\STATE Define \\$X'=x_1, \ldots, x_{i-1}, [MASK], x_i, \ldots, x_n$
		\STATE Send $X'$ into a trained BERT masked language model \citet{devlin-etal-2019-bert} by \citet{wolf-etal-2020-transformers} and get $y\in \mathbb{R}^d$, a probability vector for each word in the vocabulary (\textit{d} is the size of the vocabulary of the BERT model)
		\STATE Define $w_k\in W$ s.t. $w_k=\max_{i=1,\ldots m}y[w_i]$ to be a new word at position \textit{i} of sentence $X$ (the word with the highest probability, according to BERT, out of the given set $W$).
\end{algorithmic}
\begin{algorithmic}
\STATE{\bfseries Output:} a sentence\\$\left(x_1, x_2, \ldots, x_{i-1}, w_k, x_i, \ldots, x_n\right)$
\end{algorithmic}
We can make this an \textbf{Optional Word Insertion}\label{tool:opt_wi} by returning either the input or output sentence, using Sentence Probability.

\paragraph{Noun Derivation.}
\begin{algorithmic}
\STATE{\bfseries Input:} a verb (lemma form)
\STATE We prioritize choosing the noun form from AMR morph verbalization \footnote{\url{https://amr.isi.edu/download.html}}. If we don't find it there, we choose between Nomlex form \citep{Macleod98nomlex:a} and present participle form according to Verb Form Dictionaries\footnote{\url{https://github.com/monolithpl/verb.forms.dictionary}} (if exists), deciding according to Word Insertion.
\STATE{\bfseries Output:} either a noun or None
\end{algorithmic}

\paragraph{Preposition Sets.}
Using Word Insertion requires a set of options as input. In our paraphrasing process, we use the following predefined sets to insert prepositions.
\textbf{Temporal prepositions: }
{'as', 'aboard','along', 'around', 'at', 'during', 'upon', 'with', 'without'}.
\textbf{General prepositions:}
{'as', 'aboard', 'about', 'above', 'across', 'after', 'against', 'along', 'around',
                'at', 'before', 'behind', 'below', 'beneath', 'beside', 'between', 'beyond', 'but',
                'by', 'down', 'during', 'except', 'following', 'for', 'from', 'in', 'inside',
                'into', 'like', 'minus', 'minus', 'near', 'next', 'of', 'off', 'on', 'onto', 'onto',
                'opposite', 'out', 'outside', 'over', 'past', 'plus', 'round', 'since', 'since',
                'than', 'through', 'to', 'toward', 'under', 'underneath', 'unlike', 'until', 'up',
                'upon', 'with', 'without'}.

\subsection{Active Voice to Passive Voice}\label{conv:active}
The active-to-passive paraphrasing process is done on sentences that include a nominal subject and a direct object. We discard any sentence of question and coordination, possible passive form (root verb is in past participle), and those where the root verb has a \textit{"to"} auxiliary.


\begin{algorithmic}[1]
		\STATE If the subject is a proper noun, convert it to object form
		\STATE If the direct object is a proper noun, convert it to subject form
		\STATE Switch the subtree spans of subject and object
		\STATE Add "by" just before the span of the new object
		\STATE If an auxiliary verb is one of "can", "may", "shall", convert it to "could", "might", "should" respectively.
		\STATE If root verb is a gerund or present participle, replace it with "being". Otherwise, remove it altogether.
		\STATE Add suitable auxiliary according to the new subject form of singular/plural, and the tense.
		\STATE If the sentence includes a negation word, remove it and add "not" before the auxiliary.
		\STATE Replace the root verb to its past participle form (using the Verb Forms Dictionary\footnote{\url{https://github.com/monolithpl/verb.forms.dictionary}}).
		\STATE If the sentence includes a particle, move it after the root verb.
		\STATE If the sentence includes a dative, try to replace it using Optional Word Insertion.
\end{algorithmic}

We'll go over an example:
\paragraph{Active to Passive: example}
\begin{algorithmic}
\STATE{\bfseries Input:} He can't take the book.
\end{algorithmic}
\begin{algorithmic}[1]
		\STATE $"He" \gets "Him"$
		\STATE NA
		\STATE Switch \textit{"him"} with \textit{"The book"}
		\STATE $"him" \gets "by\;him"$
		\STATE $"can" \gets "could"$
		\STATE NA
		\STATE Add \textit{"be"}
		\STATE $"'t" \gets "not"$
		\STATE $"take" \gets "taken"$
		\STATE NA
		\STATE NA
\end{algorithmic}
\begin{algorithmic}
\STATE{\bfseries Output:} The book could not be taken by him.
\end{algorithmic}


\begin{table*}[thb]
 \begin{threeparttable}
	\centering
		\begin{tabular}{c|c|c|c|c}
			\toprule
			{\cellcolor{LightGray}} & \multicolumn{4}{c}{\textbf{Extract Adverbial Clause}}\\
			\multirow{-2}{*}{\cellcolor{LightGray}1. Extract}&\multicolumn{4}{c}{detect type by Semantic Role Labeling \citep{gardner-etal-2018-allennlp}}\\
			\midrule
			{\cellcolor{LightGray}} & \multicolumn{2}{c|}{\textbf{Cause/Reason}}& \textbf{Temporal} &\textbf{Purpose}\\
			{\cellcolor{LightGray}} & Possessive & Non-Possessive &   &  \\
			\midrule
			\cellcolor{LightGray}& & & marker "as"/"before"/ & \\
			 \cellcolor{LightGray}2. Match & root "have" and & root isn't: "have"/  & "after"/"until"/"while" & participle "to"\\
			 {\cellcolor{LightGray} pattern} & marker "because" &  "be"/"do"/"can" &  or adverbial modifier & \\
			 \cellcolor{LightGray}& & &  "when" & \\
			 \midrule 
			 \cellcolor{LightGray}3. aux& \multicolumn{4}{c} {remove root's auxiliaries}\\
			 \midrule 
			 \cellcolor{LightGray}& remove& \multicolumn{3}{c} {}\\
			 \cellcolor{LightGray}& direct object's& \multicolumn{3}{c} {}\\
			 \multirow{-3}{*}{\cellcolor{LightGray}4. det/Noun}& determinants& \multicolumn{3}{c} {\multirow{-3}{*}{Noun Derivation~\ref{sec:tools}}}\\
			 \midrule 
			 \cellcolor{LightGray}5. Possession& \multicolumn{4}{c} {Nominal subject to possessive form}\\
			 \midrule 
			 \cellcolor{LightGray}& \multicolumn{2}{c|}{replace "because"}& If "as"/"while"/ & Replace "to"\\
			 \cellcolor{LightGray}&\multicolumn{2}{c|}{with "because of"}& "when", replace by&with "for"\\
			 \multirow{-3}{*}{\cellcolor{LightGray}6. Preposition}& \multicolumn{2}{c|}{}& Word Insertion~\ref{sec:tools}\tnote{a}& \\
			 \midrule
			 \cellcolor{LightGray}& If negation,& \multicolumn{3}{|c}{If there is a direct object\tnote{b}} \\
			 \multirow{-2}{*}{\cellcolor{LightGray}7. Additions}& add "lack of"&\multicolumn{3}{|c}{Optional Word Insertion~\ref{sec:tools}\tnote{c}}\\
			\bottomrule
		\end{tabular}
	\begin{tablenotes}
        \item[a] Using temporal prepositions set.
        \item[b] If there is a direct object of the form "<xxx>self" in the non-possessive cause/reason case, we instead add "self" before the derived noun and remove this object.
        \item[c] Using general prepositions set.
    \end{tablenotes}
	\caption{The paraphrasing process from an adverbial clause sentence to a noun phrase.
		\label{tab:clause_para}}
		
 \end{threeparttable}
\end{table*}

The complete process of paraphrasing a sentence with an adverbial clause to one with a noun phrase substituting it is detailed in Table~\ref{tab:clause_para}. 
We'll demonstrate a few examples. \footnote{The flow of purpose clause conversion could arguably lack optional determiner addition before the new noun phrase. It could be easily added to the generation code for any future use.}
\paragraph{Purpose clause}
\begin{algorithmic}
\STATE{\bfseries Input:} She sat under the sun to enjoy the warmth.
\end{algorithmic}
\begin{algorithmic}[1]
		\STATE Extract \textit{"to enjoy the warmth"}
		\STATE Found matching participle \textit{"to"}
		\STATE NA
		\STATE $"enjoy" \gets "enjoyment"$
		\STATE NA
		\STATE $"to" \gets "for"$
		\STATE $"the warmth" \gets "of the warmth"$
\end{algorithmic}
\begin{algorithmic}
\STATE{\bfseries Output:} She sat under the sun for enjoyment of the warmth.
\end{algorithmic}

\paragraph{Cause/Reason clause, possessive form}
\begin{algorithmic}
\STATE{\bfseries Input:} She was at the library for a long time because she had an unresolved problem.
\end{algorithmic}
\begin{algorithmic}[1]
		\STATE Extract \textit{"because she had an unresolved problem"}
		\STATE Found matching root \textit{"had"} and a marker \textit{"because"}
		\STATE Remove \textit{"had"}
		\STATE Remove \textit{"an"}
		\STATE $"she" \gets "her"$
		\STATE $"because" \gets "because\;of"$
		\STATE NA
\end{algorithmic}
\begin{algorithmic}
\STATE{\bfseries Output:} She was at the library for a long time because of her unresolved problem.
\end{algorithmic}

\paragraph{Cause/Reason clause, non-possessive form}
\begin{algorithmic}
\STATE{\bfseries Input:} This robot is very advanced because it flies itself.
\end{algorithmic}
\begin{algorithmic}[1]
		\STATE Extract \textit{"because it flies itself"}
		\STATE Found matching root \textit{"flies"} and a marker \textit{"because"}
		\STATE NA
		\STATE $"flies" \gets "flight"$
		\STATE $"it" \gets "its"$
		\STATE $"because" \gets "because\;of"$
		\STATE $"flight" \gets "self\;flight"$
\end{algorithmic}
\begin{algorithmic}
\STATE{\bfseries Output:} This robot is very advanced because of its self flight.
\end{algorithmic}

\subsection{Filtering Results}\label{ap:filtering}
As mentioned in \S\ref{sec:dataset}, some sentences generated by our paraphrasing process are disfluent. Therefore, we manually filtered the data. Two in-house annotators made binary predictions as to whether the generated paraphrases are fluent, with 75\% observed agreement and 0.6 Cohen's kappa. We also tried using Direct Assessment \citep{graham2017can} and eliciting fluency scores through crowdsourcing, as well as attempting to threshold the probability given by GPT2 or SLOR \citep{kann-etal-2018-sentence}. Neither of these approaches worked in a satisfactory manner.

\section{Detecting Correlation Patterns}

\subsection{Pooling Techniques}\label{appendix:pooling}
\begin{figure*}[tbh]
\begin{center}
    \begin{subfigure}{.24\textwidth}
      \includegraphics[width=1\linewidth]{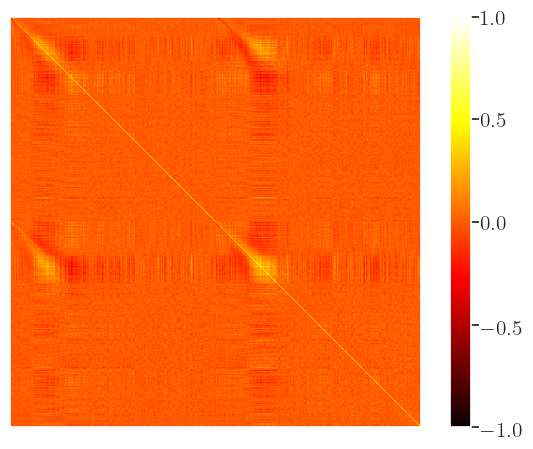}
      \caption{Mean}
      \label{fig:pool_mean}
    \end{subfigure}
    \begin{subfigure}{.24\textwidth}
      \includegraphics[width=1\linewidth]{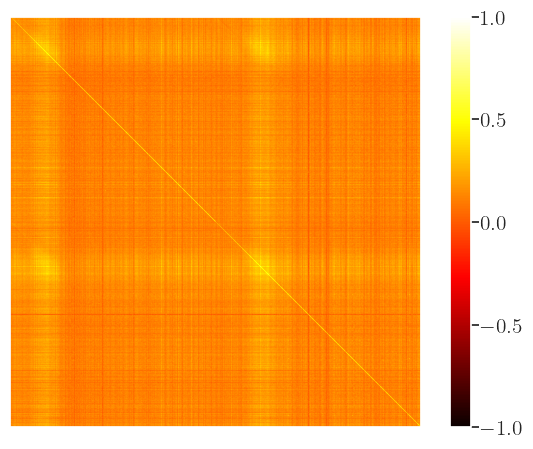}
      \caption{Min}
      \label{fig:pool_min}
    \end{subfigure}
    \begin{subfigure}{.24\textwidth}
      \includegraphics[width=1\linewidth]{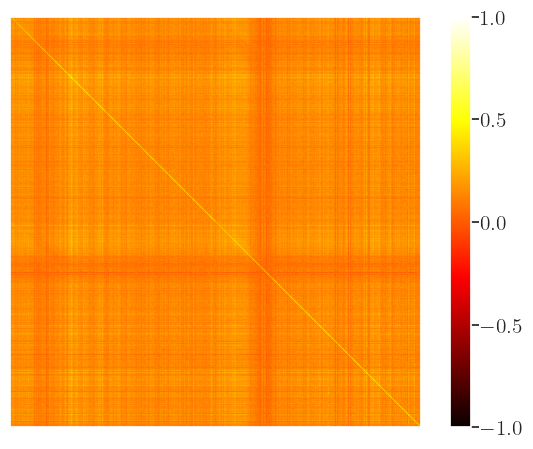}
      \caption{Max}
      \label{fig:pool_max}
    \end{subfigure}
    \end{center}
    \caption{Activation correlation between paraphrases (ParaCorr), using the active-passive dataset. The correlation is done sentence-wise, while the pooling technique of token activation varies. The major confounds are present across all techniques.}
    \label{fig:pooling}
\end{figure*}
In section \ref{sec:correlation} we measure the correlation of activations over paraphrases. Since paraphrases vary in their sequence length, the subsequent random variables representing the activations for each sentence structure (i.e. activations over active voice versus activations over passive voice), vary as well. While previous works \citep{bau2018identifying, dalvi2019one, wu-etal-2020-similarity, Meftah2021NeuralSD} compared activations over all input words, our settings necessitate pooling. In \S\ref{sec:correlation} we presented the results where we used mean activation per sentence. In Figure \ref{fig:pooling} we compare the heatmaps of different poolings. As is evident, the major confounds (diagonals and concentrated neuron groups of strong correlation) are present across the techniques.

\subsection{Inside the Layer Block}\label{appendix:inside_block}
\begin{figure*}[tbh]
\begin{center}
    \begin{subfigure}{.24\textwidth}
      \includegraphics[width=1\linewidth]{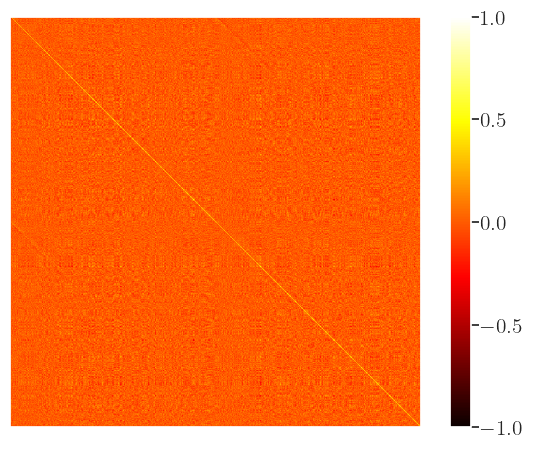}
      \caption{Post Attention}
      \label{fig:post_attn}
    \end{subfigure}
    \begin{subfigure}{.24\textwidth}
      \includegraphics[width=1\linewidth]{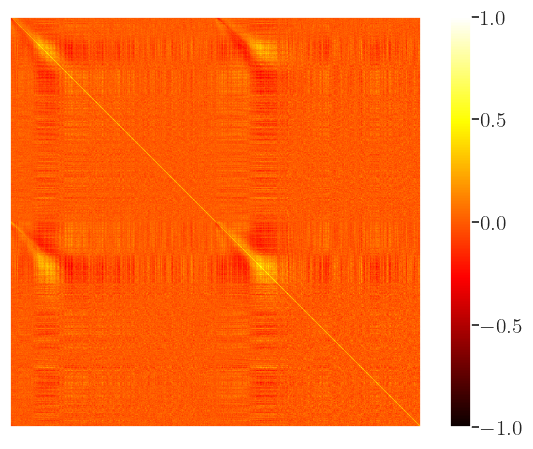}
      \caption{Residual + Norm}
      \label{fig:pre_fc}
    \end{subfigure}
    \begin{subfigure}{.24\textwidth}
      \includegraphics[width=1\linewidth]{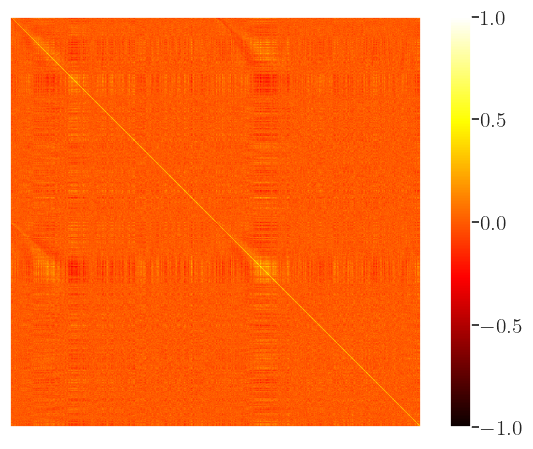}
      \caption{Linear Layers}
      \label{fig:post_fc}
    \end{subfigure}
    \begin{subfigure}{.24\textwidth}
      \includegraphics[width=1\linewidth]{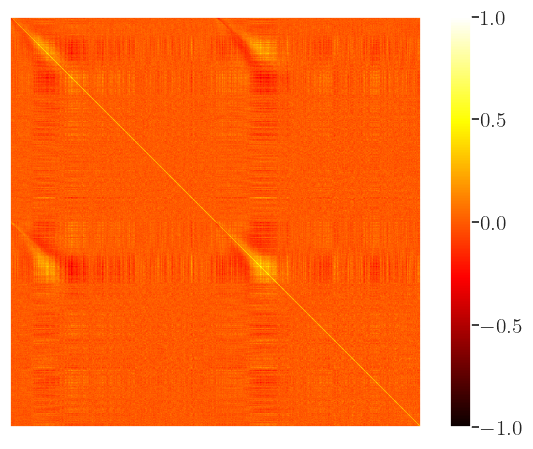}
      \caption{Residual + Norm}
      \label{fig:layer}
    \end{subfigure}
    \end{center}
    \caption{Activation correlation between paraphrases (ParaCorr), using the active-passive dataset. A view inside the first encoder layer block, step-by-step: (a)  attention heads, (b) adding residual connections and applying normalization, (c) fully connected layer, followed by ReLU and another fully connected layer, (d) adding residual connections and applying normalization - the output of the layer block.}
    \label{fig:corr_inside}
\end{figure*}
In section \ref{sec:correlation} we measure the correlation of activations only at the output of the encoder layer block, following previous work \citep{wu-etal-2020-similarity}. We also take a look at intermediate activations, see Figure \ref{fig:corr_inside}. This strengthens our hypothesis that the strong correlation seen in PosCorr (Figure \ref{fig:pos_corr}) is due to the sinusoidal positional encodings, as they are propagated through the network with residual connections. The PosCorr effect appears only after the first residual connection, weakens through the fully-connected layers, and strengthens again after additional residual connection.

\subsection{Adverbial Clause versus Noun Phrase}\label{appendix:corr_clause}
\begin{figure*}[tbh]
\begin{center}
    \begin{subfigure}{.24\textwidth}
      \includegraphics[width=1\linewidth]{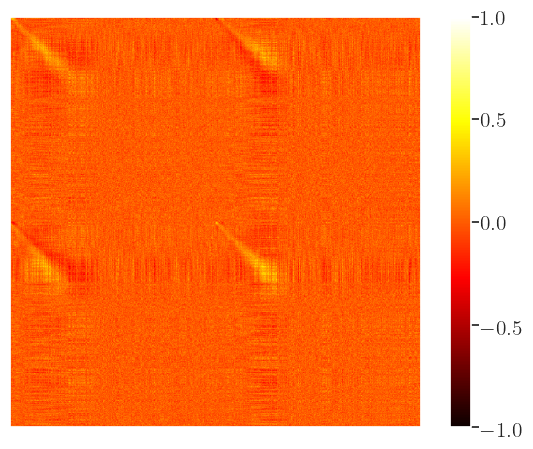}
      \caption{ModelCorr}
      \label{fig:clause_model_corr}
    \end{subfigure}
    \begin{subfigure}{.24\textwidth}
      \includegraphics[width=1\linewidth]{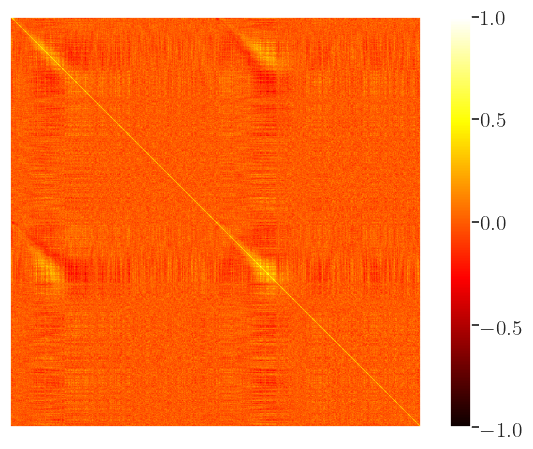}
      \caption{ParaCorr}
      \label{fig:clause_para_corr}
    \end{subfigure}
    \begin{subfigure}{.24\textwidth}
      \includegraphics[width=1\linewidth]{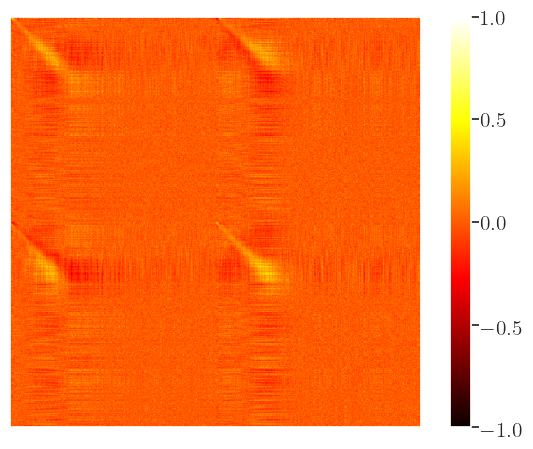}
      \caption{PosCorr}
      \label{fig:clause_pos_corr}
    \end{subfigure}
    \begin{subfigure}{.24\textwidth}
      \includegraphics[width=1\linewidth]{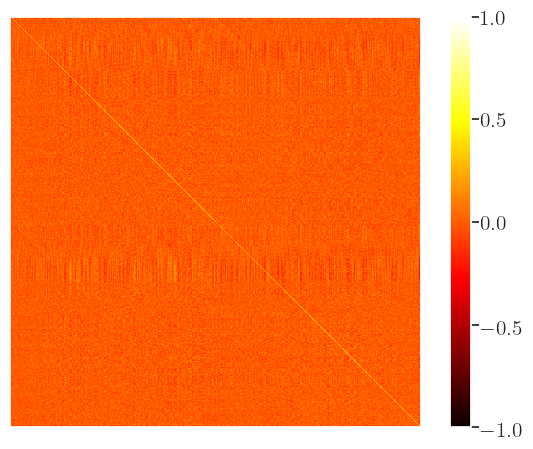}
      \caption{TokenCorr}
      \label{fig:clause_token_corr}
    \end{subfigure}
    \end{center}
    \caption{Activation correlation of first layer neurons in the Transformer encoder, using the clause/noun-phrase dataset.}
    \label{fig:clause_corr_maps}
\end{figure*}
Here we present the same correlation methods detailed in section \ref{sec:correlation} but measured on the adverbial clause versus noun phrase sets. See Figure \ref{fig:clause_corr_maps}.

\section{Manipulation of Neurons}\label{appendix:manipulation}

\subsection{Active to Passive}
\begin{figure*}[tbh]
    \begin{subfigure}{.33\textwidth}
      \centering
      \includegraphics[width=1\linewidth]{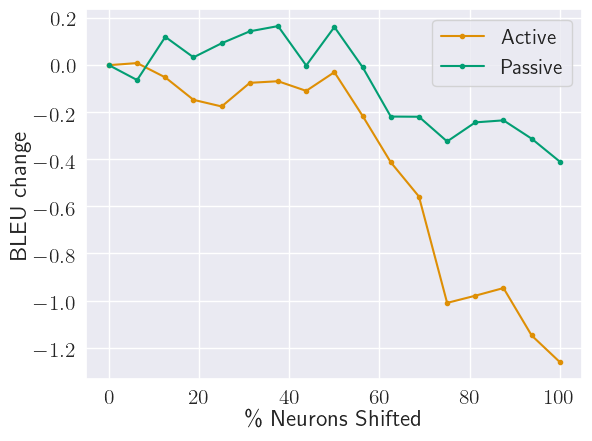}
      \caption{Baseline}
      \label{fig:active_baseline}
    \end{subfigure}
    \begin{subfigure}{.33\textwidth}
      \centering
      \includegraphics[width=1\linewidth]{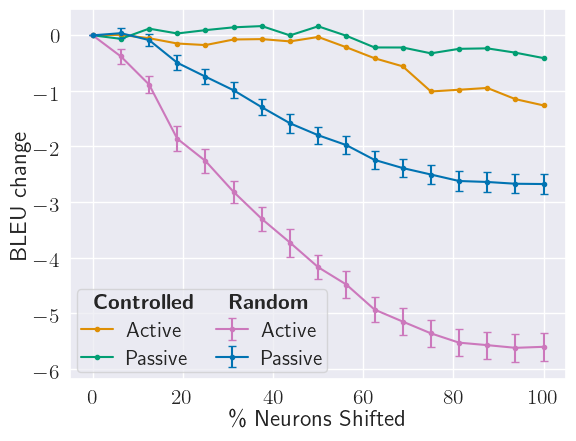}
      \caption{Random direction comparison}
      \label{fig:active_randdir}
    \end{subfigure}
    \begin{subfigure}{.33\textwidth}
      \centering
      \includegraphics[width=1\linewidth]{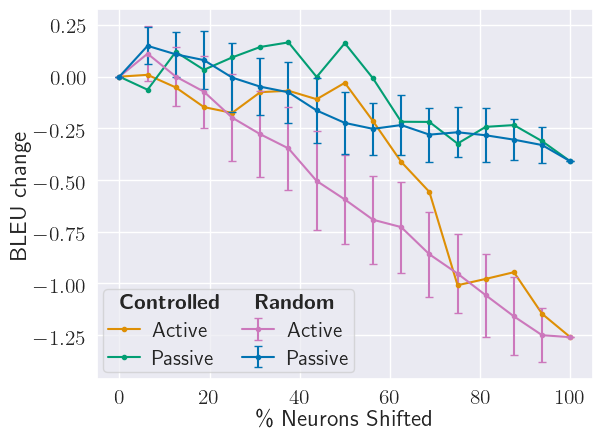}
      \caption{Random selection comparison}
      \label{fig:active_randselect}
    \end{subfigure}
    
    \caption{Manipulating output translation to be in passive voice when feeding active voice as input. Lines present BLEU change with active and passive references according to the amount of neurons manipulated (x).}
    \label{fig:ap_shift_bleu_active}
\end{figure*}
To complete all variations of the manipulation experiment, we first showcase the shift from active voice input to passive voice translation (the opposite direction of what we showed in the main paper). We see that the translation is more similar to the target form (passive voice) than the input form (active voice). The positive change in BLEU is more subtle in this manipulation, and again getting maximal change requires many neurons to be modified (at least 50\%), see Figure \ref{fig:active_baseline}. With the random experiments of direction (Figure \ref{fig:active_randdir}) and neurons selection (Figure \ref{fig:active_randselect}), we get similar results - our controlled direction is better while choosing an optimal subset of neurons is not easy.

\subsection{Manipulation on a Test Set}\label{ap:test_set}
\begin{figure}[tbh]
      \centering
      \includegraphics[width=.33\textwidth]{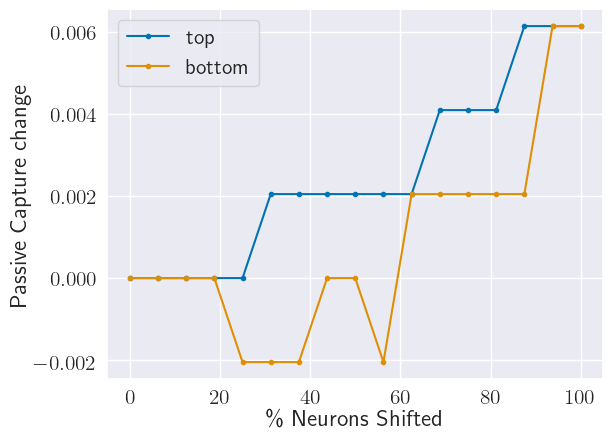}
    \caption{Manipulating neurons to get passive voice translation given an active voice input from the test set. Comparing the effect of manipulating first top versus bottom neurons, according to ParaCorr. We measure passive form detection}
    \label{fig:test_shift_score}
\end{figure}

We repeat the manipulation on a held-out test set: 552 sentences that we automatically detect as active voice sentences from the WMT19 test set. While our experiments on the dev set are valid, as we manipulate from one set (e.g. passive voice) by measuring another (e.g. active voice), one might argue that we can't know the effect of the shared semantic meaning (on the set level) has on the success rate. To cover all bases, we manipulate the test set according to average activations measured on the dev set. Here we do not have a passive voice counterpart, so we manipulate active voice inputs to passive voice translations.
The passive voice detection score (see \S\ref{sec:beyond_bleu}) shows a monotonous increase (up to 0.6\% more) as we modify more neurons (see Figure \ref{fig:test_shift_score}). The trend matches our expectations. Moreover, we see again that manipulating top-ranked neurons (rank given by ParaCorr) has a greater effect than bottom-ranked ones. This is again consistent with what we saw with the development set and BLEU score in section \ref{sec:man_tb}.

\begin{figure}[tbh]
\centering
    \begin{subfigure}{.22\textwidth}
      \centering
      \includegraphics[width=1\linewidth]{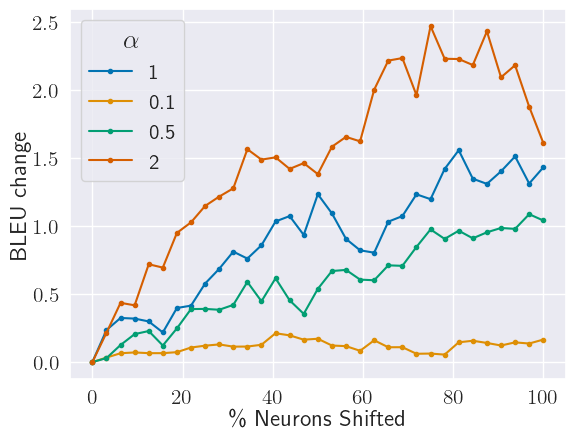}
      \caption{Passive to Active}
      \label{fig:passive2active_grid}
    \end{subfigure}
    \begin{subfigure}{.22\textwidth}
      \centering
      \includegraphics[width=1\linewidth]{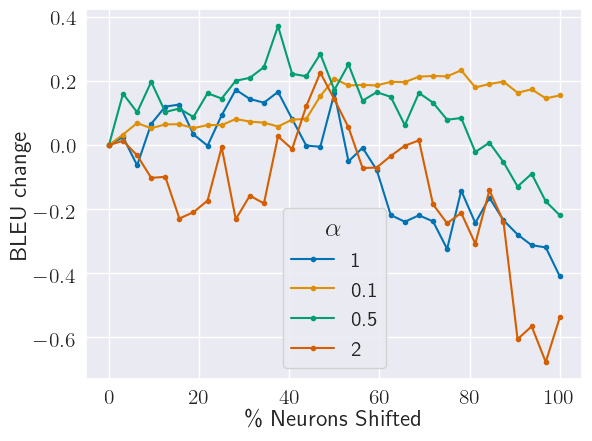}
      \caption{Active to Passive}
      \label{fig:active2passive_grid}
    \end{subfigure}
    \caption{Comparing various magnitudes $\alpha$ for manipulation step $\frac{\alpha}{\|{\bar{x_{c_1}}-\bar{x_{c_2}}}\|}(\bar{x_{c_1}}-\bar{x_{c_2}})$. BLEU score measured against reference of target form, when manipulating increasingly more neurons according to top rank of ParaCorr.}
    \label{fig:grid_ap}
\end{figure}

\subsection{Noun Phrase to Adverbial Clause}
\begin{figure*}[tbh]
    \begin{subfigure}{.33\textwidth}
      \centering
      \includegraphics[width=1\linewidth]{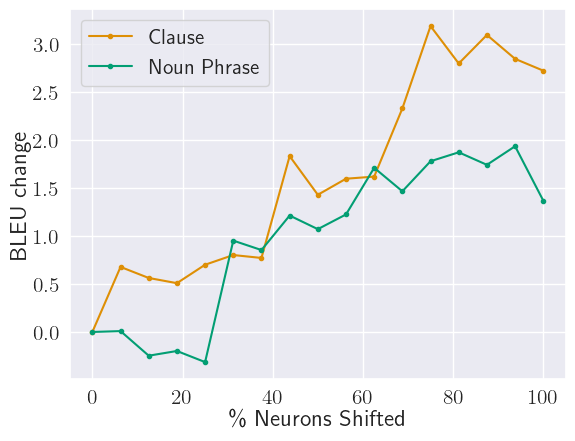}
      \caption{Baseline}
      \label{fig:nphrase2clause_baseline}
    \end{subfigure}
    \begin{subfigure}{.33\textwidth}
      \centering
      \includegraphics[width=1\linewidth]{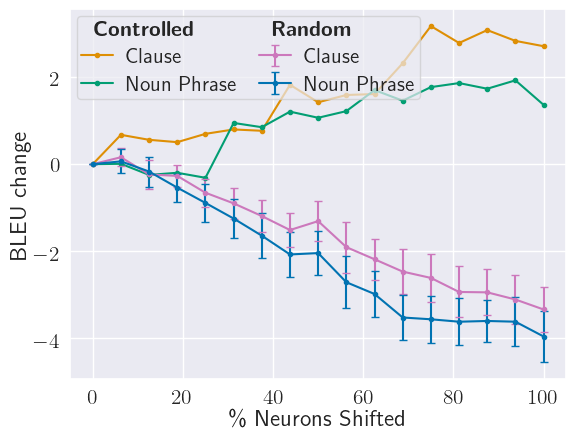}
      \caption{Random direction comparison}
      \label{fig:nphrase2clause_randdir}
    \end{subfigure}
    \begin{subfigure}{.33\textwidth}
      \centering
      \includegraphics[width=1\linewidth]{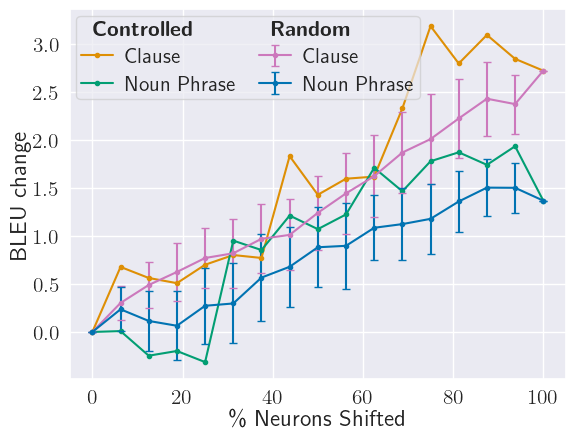}
      \caption{Random selection comparison}
      \label{fig:nphrase2clause_randselect}
    \end{subfigure}
    
    \caption{Manipulating output translation to be with an adverbial clause when feeding a sentence with a noun phrase as input. Lines present BLEU change with active and passive references according to the amount of neurons manipulated (x).}
    \label{fig:nphrase2clause_shift_bleu}
\end{figure*}

Manipulating from a noun phrase to an adverbial clause is consistent with the results we saw for passive to active manipulation, see Figure \ref{fig:nphrase2clause_shift_bleu} 
We repeat the same succession of experiments on the adverbial clause versus noun phrase dataset. 

\subsection{Adverbial Clause to Noun Phrase}\label{sec:clause2nphrase_diss}
\begin{figure*}[tbh]
    \begin{subfigure}{.33\textwidth}
      \centering
      \includegraphics[width=1\linewidth]{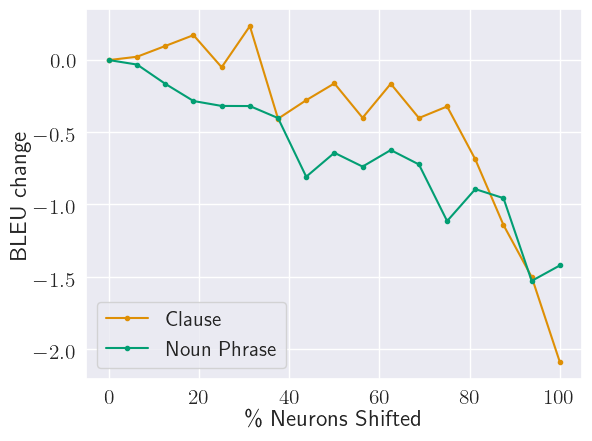}
      \caption{Baseline}
    \end{subfigure}
    \begin{subfigure}{.33\textwidth}
      \centering
      \includegraphics[width=1\linewidth]{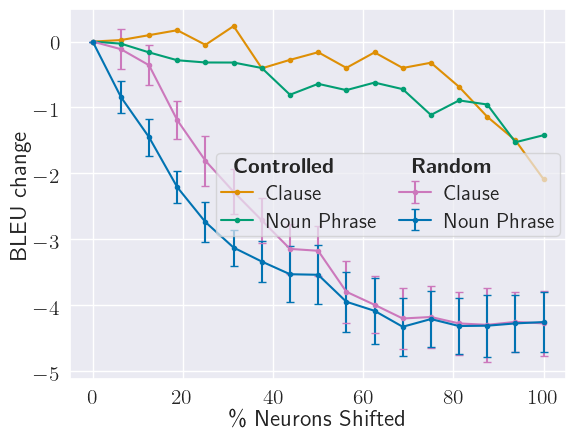}
      \caption{Random direction comparison}
      \label{fig:clause2nphrase_randdir}
    \end{subfigure}
    \begin{subfigure}{.33\textwidth}
      \centering
      \includegraphics[width=1\linewidth]{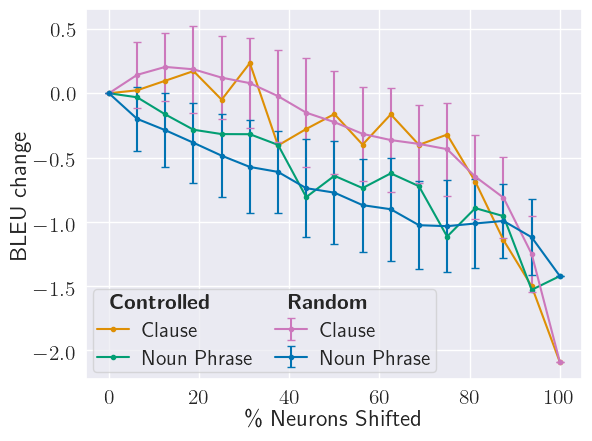}
      \caption{Random selection comparison}
      \label{fig:clause2nphrase_randselect}
    \end{subfigure}
    
    \caption{Manipulating output translation to be with a noun phrase when feeding a sentence with an adverbial clause as input. Lines present BLEU change with active and passive references according to the amount of neurons manipulated (x).}
    \label{fig:clause2nphrase_shift_bleu}
\end{figure*}

Manipulating neurons to convert input with an adverbial clause to output translation with a noun phrase is not outright successful (see Figure \ref{fig:clause2nphrase_shift_bleu}). In the controlled case (where we employ direction by our records of average activation of each paraphrase form and select an increasing set of neurons to manipulate according to the top or bottom ParaCorr rank), we are still closer to the clause form than the noun phrase. We propose several possible explanations:
\begin{enumerate}
    \item The clause versus noun phrase dataset is substantially smaller than the active versus passive one (114 examples compared to 1,169 instances). A small dataset may include more noise or simply make the target syntactic form harder to capture.
    \item Adverbial clause form may be more common in the train set so the model regularizes to the statistically more acceptable option. We see hints for that when we compare the manipulation towards active form as more successful than passive form (\S\ref{sec:manipulation} and Figure \ref{fig:ap_shift_bleu_active}).
    \item Noun phrase form may not be distinctive enough to be encoded in the model.
    \item The target form may not be natural in the target language. As we discuss in our qualitative analysis in section \ref{sec:beyond_bleu}, fail cases revealed instances where the target form was either not possible for a native German speaker, or required a replacement of the verb to a synonym. This replacement demands another level of manipulation from the model, one that it may not even know to generalize.
\end{enumerate}

\subsection{Manipulation Magnitude}\label{appendix:magnitude}
As defined in section \ref{sec:manipulation}, manipulation from sentence feature $c_1$ to $c_2$ is a subtraction of the term $\frac{\alpha}{\|{\bar{x_{c_1}}-\bar{x_{c_2}}}\|}(\bar{x_{c_1}}-\bar{x_{c_2}})$ (applied to chosen neurons).
We experimented with a small grid search for possible values for the scaling factor $\alpha$, without an apparent option being better than the baseline ($\alpha=1$). See Figure \ref{fig:grid_ap} for results\footnote{We experimented with even greater values ($\alpha\in\{5, 10, 100, 1000\}$), each with a more drastic BLEU drop, therefore we discard their inclusion in the figure to allow the y-axis range to capture the subtle trends of the variables presented.}. There is no definitive conclusion of what magnitude would be consistently better in every manipulation. Similar trends were found in the clause dataset: $\alpha=2$ was best when manipulating from paraphrased form noun phrase back to the original form of the adverbial clause, and worse the other way around. This could be tied to the general effect we discuss in \S\ref{sec:clause2nphrase_diss} where one direction of manipulation is more effective: changing from paraphrased form to original form. This should be further investigated in future work.

\subsection{Unparalleled Sentences Manipulation}\label{appendix:splitdata}
As seen in \S\ref{sec:man_tb}, top ParaCorr neurons were better for manipulation than those from the bottom of the rank. One possible explanation we introduced was the fact that many of those top ParaCorr neurons are also top PosCorr and TokenCorr neurons. Therefore, the effectiveness might be derived from the role polysemy of these neurons, especially when the paraphrasing calls for a change of word order (e.g. active-passive requires subject and object swap) or token identity (e.g. clause to noun phrase requires a transformation between verb and noun). This is true for cases where the paraphrases are parallel pairs, therefore they share those shallow features (tokens and word order).

\begin{figure}[tbh]
\centering
    \begin{subfigure}{.22\textwidth}
      \centering
      \includegraphics[width=1\linewidth]{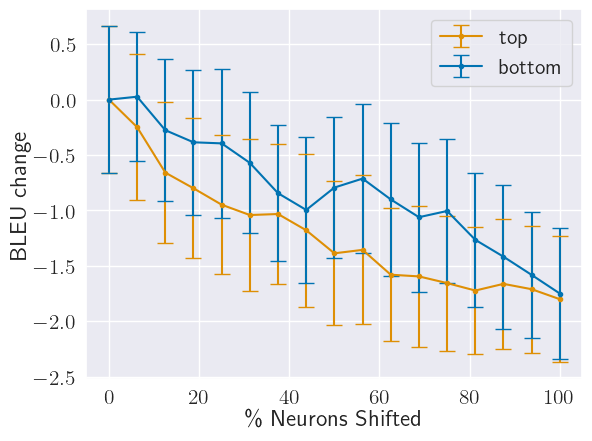}
      \caption{Passive to Active}
      \label{fig:passive2active_split}
    \end{subfigure}
    \begin{subfigure}{.22\textwidth}
      \centering
      \includegraphics[width=1\linewidth]{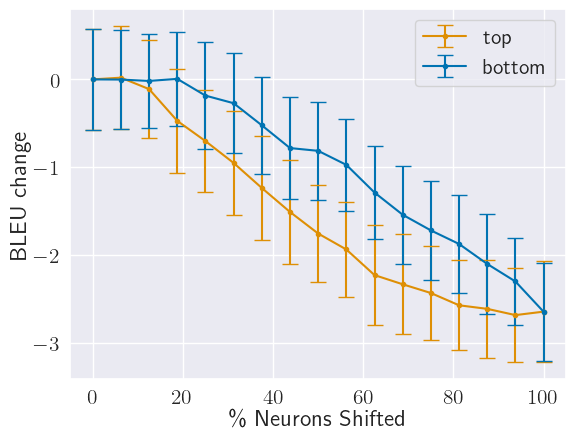}
      \caption{Active to Passive}
      \label{fig:active2passive_split}
    \end{subfigure}
    \caption{Comparing the impact of manipulating top ParaCorr neurons versus bottom ones, where the modification value of neurons is determined by average activation under unparalleled sets of sentences, i.e. $\bar{x_{c_1}}$ and $\bar{x_{c_2}}$ are measured by active and passive non-pairs. The lines represent the mean and standard deviation over 100 different unparalleled sets.}
    \label{fig:splitdata}
\end{figure}

Here we present an experiment of manipulation where there are no parallel pairs of paraphrases. We randomly split the sentence pairs into two sets. From one we take only the active sentences, and from the other we take only the passive sentences, resulting in an active set and passive set of unrelated sentences. We repeat the manipulation experiment as detailed in \S\ref{sec:man_exp} but use those unparalleled sets for measuring the averaging activation of neurons under the active voice feature and under the passive voice feature (i.e., for measuring $\bar{x_{c_1}}$ and $\bar{x_{c_2}}$). We do so for 100 different splits of the data. Measuring the mean and standard deviation of the BLEU against the objective reference, the results are presented in Figure \ref{fig:splitdata}. Notably, the standard deviation of the experiment is reported, not the standard error of the mean. The results may match the intuition where the least correlated neurons between paraphrases are those most sensitive to the active-passive feature, but since nothing is shared across those sentences, the expected noise level is high, and any measure is hard to explain.

When we measured the correlation of such unparalleled sentences, we got an average correlation (per neuron over 100 different splits of the dataset into unparalleled sets) ranging from $-0.04$ to $0.04$, with a standard deviation between $0.03$ to $0.06$. 

\section{Qualitative Analysis of Manipulation}\label{ap:quality}

Sentence examples of successful manipulation from passive voice input to active voice translation, as examined by a native German speaker, can be found in Table~\ref{tab:pa_exm}.
\begin{table*}[]
\begin{footnotesize}
\caption{Example of successfully manipulated sentences, from passive voice input to active voice translation. Manipulation is done by shifting the values of all the neurons in the encoder toward their average activation on active voice sentences. Correctness of sentence voice and fluency was verified by a native German speaker.}
\label{tab:pa_exm}
\begin{center}
\begin{tabular}{@{}lll@{}}
\toprule
\begin{tabular}[c]{@{}l@{}}Input sentence:\\passive voice\end{tabular} &
\begin{tabular}[c]{@{}l@{}}Baseline translation:\\passive voice\end{tabular} &
\begin{tabular}[c]{@{}l@{}}Manipulated translation:\\active voice\end{tabular} 
\\ \midrule
\begin{tabular}[c]{@{}l@{}}The scene was described by\\ police as very gruesome.\end{tabular}                                        & \begin{tabular}[c]{@{}l@{}}Der Tatort wurde von der Polizei\\ als sehr grauenvoll beschrieben.\end{tabular}                            & \begin{tabular}[c]{@{}l@{}}Die Polizei beschrieb den\\ Tatort als sehr grauenvoll.\end{tabular}                         \\ \\
\begin{tabular}[c]{@{}l@{}}During the excavations, the\\ remains of a total of five creatures\\ were collected by them.\end{tabular} & \begin{tabular}[c]{@{}l@{}}Bei den Ausgrabungen wurden von\\ ihnen die \"Uberreste von insgesamt\\ f\"unf Lebewesen gefunden.\end{tabular} & \begin{tabular}[c]{@{}l@{}}Bei den Ausgrabungen\\ fanden sie die \"Uberreste von\\ insgesamt f\"unf Lebewesen.\end{tabular} \\ \\
\begin{tabular}[c]{@{}l@{}}From "dream" to "megalomania":\\ the Bit Galerie is discussed by\\ TV readers\end{tabular}                & \begin{tabular}[c]{@{}l@{}}Vom "Traum" zum "Gr\"o\ss{}enwahn":\\ Die Bit-Galerie wird von TV-Lesern\\ diskutiert\end{tabular}                & \begin{tabular}[c]{@{}l@{}}Vom "Traum" zum\\ "Gr\"o\ss{}enwahn": TV-Leser\\ diskutieren \"uber die Bit-Galerie\end{tabular}     \\ \bottomrule
\end{tabular}
\end{center}
\end{footnotesize}

\end{table*}

As we discuss in \S\ref{sec:beyond_bleu}, sometimes a manipulation is not applicable in the target language. For example, the adverbial clause sentence from our dataset \textit{"In Lyman's case, she reported the alleged rape to military police less than an hour after it occurred."}, is translated into a noun phrase sentence regardless of input form (i.e. if we insert either this as input or its noun phrase paraphrase) or manipulation (i.e. with or without manipulation). "it occurred" is immediately translated into the German parallel of "its occurrence" when translating the clause version, and it is translated into a wrong noun phrase when translating the noun phrase version (the German parallel of "appearance" rather than "occurrence" in this context, i.e. "Auftreten" and "Vorfall", respectively). A native German speaker suggested we opt to replace "occurred" with "happened", otherwise it could not be translated into a clause form. Even the human reference (of WMT) is with the "its occurrence" noun phrase. See Table~\ref{tab:clause_fail}.

\begin{table*}[]
\begin{footnotesize}
\caption{Example of an adverbial clause and a noun phrase translations, showcasing the limitations of BLEU comparison to Google Translate references and the challenge of translating an output in adverbial clause form. Either manipulation here did not have any effect (e.g. manipulation from clausal input resulted in translation identical to the one without manipulation)}
\centering
\begin{tabular}{@{}lll@{}}
\toprule
English                                                    & \begin{tabular}[c]{@{}l@{}}\textbf{Adverbial Clause}\\ In Lyman's case, she reported the alleged\\ rape to military police less than an hour\\ after it occurred.\end{tabular}  & \begin{tabular}[c]{@{}l@{}}\textbf{Noun Phrase}\\ In Lyman's case, she reported the alleged\\ rape to military police less than an hour\\ after its occurrence.\end{tabular}                 \\ \\  \midrule
\begin{tabular}[c]{@{}l@{}}Human\\ Reference\end{tabular}  & \begin{tabular}[c]{@{}l@{}}In Lymans Fall meldete sie die mutma\"ssliche\\ Vergewaltigung der Milit\"arpolizei weniger als\\ eine Stunde nach dem \"uberfall. \end{tabular}    &                                                                                                                                                                                     \\ \\ 
\begin{tabular}[c]{@{}l@{}}Google\\ Translate\end{tabular} & \begin{tabular}[c]{@{}l@{}}In Lymans Fall meldete sie die mutma\"ssliche\\ Vergewaltigung weniger als eine Stunde nach\\ ihrem Auftreten der Milit\"arpolizei.\end{tabular} & \begin{tabular}[c]{@{}l@{}}In Lymans Fall wurde die mutma\"ssliche\\ Vergewaltigung von ihr weniger als eine\\ Stunde nach ihrem Auftreten der Milit\"arpolizei\\ gemeldet.\end{tabular} \\ \\ 
\begin{tabular}[c]{@{}l@{}}Our\\ Translation\end{tabular}  & \begin{tabular}[c]{@{}l@{}}In Lymans Fall meldete sie die angebliche\\ Vergewaltigung weniger als eine Stunde nach\\ dem Vorfall der Milit\"arpolizei.\end{tabular}      & \begin{tabular}[c]{@{}l@{}}In Lymans Fall meldete sie die angebliche\\ Vergewaltigung weniger als eine Stunde nach\\ ihrem Auftreten der Milit\"arpolizei.\end{tabular}              
\end{tabular}

\label{tab:clause_fail}
\end{footnotesize}
\end{table*}

\end{document}